%% file: ICAAN.tex
\documentclass{article}
\usepackage[final,nonatbib]{neurips_2018}

\usepackage[utf8]{inputenc} 
\usepackage[T1]{fontenc}    
\usepackage{hyperref}       
\usepackage{url}            
\usepackage{booktabs}       
\usepackage{amsfonts}       
\usepackage{nicefrac}       
\usepackage{microtype}      
\usepackage{amsmath,amssymb}
\usepackage[linesnumbered,boxed]{algorithm2e}
\usepackage{algorithmic}
\usepackage{graphicx}
\usepackage{epstopdf}

\input{command_templates}

\SetKwInput{KwInput}{Input}
\SetKwInput{KwOutput}{Output}
\SetKwInput{KwTask}{Task}
\SetKwInput{KwInit}{Initialization}
\SetKwInput{KwProc}{Loop}
\SetKwInput{KwStepk}{Step K}

\title{Non-Negative Kernel Sparse Coding \\ for the Classification of Motion Data}

\author{%
	Babak Hosseini
	\thanks{
		Preprint of the publication~\cite{hosseini2016non}, as provided by the authors.
		The final publication is available at Springer's Lecture Notes in Computer Science series via \url{https://link.springer.com/chapter/10.1007/978-3-319-44781-0_60}  	
	} \\
	CITEC cluster of excellence\\
	Bielefeld University, Germany\\
	\texttt{bhosseini@techfak.uni-bielefeld.de} \\
\And
Felix H\"ulsmann\\
CITEC cluster of excellence\\
Bielefeld University, Germany\\
\texttt{fhuelsma@techfak.uni-bielefeld.de} \\
\And
Mario Botsch\\
CITEC cluster of excellence\\
Bielefeld University, Germany\\
\texttt{botsch@techfak.uni-bielefeld.de} \\
	\And
	Barbara Hammer\\
	CITEC cluster of excellence\\
	Bielefeld University, Germany\\
	\texttt{bhammer@techfak.uni-bielefeld.de} \\
}

\begin{document}
\maketitle

\begin{abstract}
We are interested in the decomposition of motion data into a sparse linear combination of base functions which enable efficient data processing. 
We combine two prominent frameworks: dynamic time warping (DTW), which offers particularly successful 
pairwise motion data comparison, and sparse coding (SC), 
which enables an automatic decomposition of vectorial data into a sparse linear combination of base vectors.
We enhance SC as follows: an efficient kernelization which extends its application domain to general similarity data such as offered by DTW, and its restriction to non-negative linear representations of signals and base vectors in order to guarantee a meaningful dictionary.
Empirical evaluations on motion capture benchmarks show the effectiveness of our framework regarding interpretation and discrimination concerns.
\textit{\textbf{keywords: }}{Kernel sparse coding, motion analysis, classification, interpretable models, dynamic time warping}
\end{abstract}

\input{intro}
\input{NNKSC}
\input{experiments}
\input{conclusion}
\section{Acknowledgment}
This research was supported by the Cluster of Excellence Cognitive 
Interaction Technology 'CITEC' (EXC 277) at Bielefeld University, which
is funded by the German Research Foundation (DFG).

\bibliographystyle{unsrt}
\bibliography{C:/Thesis/Publications/Ref4Papers_CS}

\end{document}

%% file: command_templates.tex
\newcommand{\K}{\mathcal{K}}
\newcommand{\MR}[1]{\mathrm{#1}}
\newcommand{\MB}[1]{\mathbf{#1}}
\newcommand{\MBBR}{\mathbb{R}}
\newcommand{\MC}[1]{\mathcal{#1}}

%% file: intro.tex
\section {Introduction}
Ubiquitous sensors such as Microsoft's Kinect, video cameras, and
motion capturing or tracking systems cause an increasing availability of digital signals, which
describe some form of human motion data. Unless such data are
manually labeled,  its content 
is often not clear, and it remains a challenge on how to automate
semantic search in motion databases. 
In this contribution, we investigate in how far natural priors such as sparsity allow
automatic extraction of semantically meaningful entities based on the
given data alone. 

We hypothesize that semantics is mirrored by recurring
signals, which are present in semantically similar motion data, and it is
possible to infer such signals from given data based on their
property that  they allow a particularly efficient
description of the signals.
We rely on two techniques which have proven successful in such settings:
dynamic time warping enables an efficient grouping of
time series according to their semantic similarity, as has been shown
in numerous applications such as semantic classification based on the DTW distance \cite{shokoohigeneralizing}.
We rely on DTW as an interface to represent time series of possibly different length
in terms of their pairwise similarity. DTW alone does not provide
a compact representation of the given data. For the latter, we use sparse coding (SC),
which extracts a dictionary from a given data set, such that it enables
a sparse linear representation of the signals
\cite{Aharon2006}. The resulting dictionary elements constitute
an interface based on which semantic search becomes possible: signals which
decompose into the same dictionary elements have considerable semantic overlap.

The combination of DTW and SC faces two problems: 
SC deals with vectorial data; hence, we resort to a kernel version
of SC
\cite{chen2015kernel}.
Besides, negative coefficients provide unreasonable base functions by linear combinations.
Therefore, we extend kernel SC to a non-negative version.
We demonstrate the accuracy of the proposed method for 
various motion capture benchmarks.

%% file: NNKSC.tex
\section {None Negative Kernel Sparse Coding}
Given a measurement signal $\vec y \in \MBBR^{n}$ which is an element of a set of measurements $\MB{Y} \in \MBBR^{n\times N}$, sparse
coding finds a representation $\vec y = \MB{D}\vec x$ of the signal $\vec y$. The Dictionary $\MB{D}\in \MBBR^{n\times k}$ constitutes a matrix of basic primitives
which are shared by all measurements in $\MB{Y}$, and $\vec x$ constitutes a sparse coefficient vector which
describes how the observation $\vec y$ is generated by the basic primitives. 
Considering the motion data in the vectorial space, each data exemplar $\MB{Y}_i$ from the training set $\MC{Y}=\{\MB{Y}_i\}_{i=1}^N$ would be a time-series matrix $\MB{Y}_i=[\vec{y}_i(1). . .\vec{y}_i(T)] \in (\mathbb{R}^n)^\ast$ and potentially has a different length than other exemplars. Therefore, to facilitate the analysis, we use the distance between the data samples, and we form the kernel function $\K(\MB{Y}_i,\MB{Y}_j)$. 

By using the kernel function, the sparse coding problem would become the following estimation $\Phi(\MB{Y}_i) = \Phi(\MB{D})\vec x_i$. The dictionary matrix $\Phi(\MB{D})$ is defined in the feature space, and the underlying kernel function is generally not available, and so it is difficult to estimate it in this form. Hence, we use the trick suggested by \cite{VanNguyen2013}, as defining $\Phi(\MB{D})=\Phi(\MB{Y})\MB{A}$ that gives us the opportunity to update the dictionary in the input space via matrix $\MB{A}\in \mathbb{R}^{N \times k}$. 

From another point of view, As we are dealing with motion data, we'd like to have a dictionary such that its elements have the characteristics of motion signals. That way, they can be considered as motion representatives or the prototypes for different motion groups. Furthermore, the resulting model can be linked to semantically meaningful entities, such that each motion sample can be related to one or more meaningful motion prototypes. One way to achieve this goal is to formulate the SC problem such that the motion signals would be coded with non-negative coefficient vectors as $\vec{x} \ge 0$. 

Furthermore, we like to determine dictionary $\MB{D}$ such that its representative matrix $\MB{A}$ uses as few 
elements from $\Phi(\mathbf Y)$ as possible (small $\| \vec a_j \|_1$), which can result in a sparser dictionary. In other words, it leads to using fewer signals from the whole set of $\mathbf Y$ for representing the motion dataset. Also as we are modeling motion signals, a dictionary atom which is the positive linear combination of input data ($a_{ij} \ge 0$) would have higher chances to be semantically meaningful in the context of motion data. As a result, we can formulate our desired non-negative kernel sparse coding framework as: 
\begin{equation}
\begin{array}{ll}
\underset{\MB{X},\MB{D}}{\mathrm{min}}  &\|  \Phi({\MC{Y}})-\Phi({\MC{Y}}){\mathbf A}{\mathbf X}\|_F^2 +\|\MB{A}\|_1^2 \\
\mathrm{s.t} &\|\vec x_i\|_0\le T , ~\forall i=1...N. , \quad a_{ij} \ge 0 , \quad \ x_{ij} \ge 0 \\
\label{eq:NNkksvd}
\end{array}
\end{equation}
In order to solve the optimization problem in (Eq.\ref{eq:NNkksvd}), we use alternating optimization based on the two proposed kernel based optimization algorithms as ``Non-Negative Kernel Orthogonal Matching Pursuit (NN-KOMP)'' and ``Non-Negative Kernel dictionary learning''. These methods estimate the sparse coefficients $\MB{X}$ and the dictionary representative matrix $\MB{A}$ respectively.
The code of NN-KSC and its supervised version (LC-NNKSC) are available from the public online repository\footnote{https://github.com/bab-git/NNKSC}.

\subsection{Non-Negative Kernel OMP}\label{sec:NN-KOMP}
For the sparse coding part of Eq.\ref{eq:NNkksvd}, we want to estimate the non-negative sparse vectors $\mathbf x_i$ based on the current value of the dictionary coefficient $\MB{A}$ in order to reconstruct each motion signal $\MB{Y}_i$ as in Eq.\ref{eq:knnomp}. To that aim, we propose the NN-KOMP algorithm (Alg.\ref{fig:nnkomp}) which is based on the KOMP algorithm in \cite{VanNguyen2013}. 
However, in the first step of NN-KOMP, we select only dictionary atoms with positive correlation to the remaining residual error in each step, and in step 4, we estimate the corresponding vector $\vec x_I$ to the currently selected dictionary atoms $\MB{A}_I$ via kernelizing the Non-negative least square algorithm (K-NNLS).
\begin{equation}
\begin{array}{ll}
\vec x_i=\underset{\vec x}{\mathrm{\arg min}}  &\|  \Phi(\MB{Y}_i)-\Phi({\MC{Y}}){\mathbf A}\vec x\|_2^2 \quad \mathrm{s.t} \quad x_i \ge 0 , \quad \|\vec x\|_0\le T\\
\label{eq:knnomp}
\end{array}
\end{equation}

For the K-NNLS method, we use the active set ''lsqnonneg'' optimization algorithm from \cite{shure2006brief}, and we kernelize the necessary parts as in Algorithm \ref{fig:knnls}. As a result, the output of the K-NNLS would be used as the solution $\vec x$ in step 4 of the NN-KOMP algorithm.

	\begin{algorithm}[!t]
		\caption{The NN-KOMP algorithm}
		\label{fig:nnkomp}	
		\KwInput{Data sample $\MB{Y}$, dictionary matrix $\MB{A}$, sparseness limit T, kernel $\K$}
		\KwOutput{Approximate solution $\vec x$ to Eq.\ref{eq:knnomp}}
		\KwInit{$\vec x=0,I=\emptyset$} 
		\KwProc{}
		\Indp
		$\tau_i=max([\K(\MB{Y},\MC{Y})-\vec x_I^\top \MB{A}_{I}^\top \K(\MC{Y},\MC{Y})] \vec a_i ~,0), ~ \forall i \not\in I$ \;
		$i_{max}=\MR{arg~max}_i \lvert \tau_i \rvert, ~ \forall i \not\in I$ \;
		$I=I \cup i_{max}$ \;
		Solving $\mathrm{min_{\vec{x}}}\| \Phi(\MB{Y})-\Phi({\MC{Y}}){\mathbf A}_I \vec x\|_2^2$ ~s.t $x_i \ge 0$ ~ using K-NNLS\;
		Stop if $\| \vec x \|_0=T$\;
\end{algorithm}
	

	\begin{algorithm}[!t]
		\SetAlgoLined		
		\KwInput{Data sample $\MB{Y}$, subset matrix $\MB{A}_I \in \mathbb{R}^{n \times k}$, kernel $\K$}
		\KwOutput{Solution $\vec x$ to step 4 of NN-KOMP (Alg.\ref{fig:nnkomp})}
		\KwInit{$\vec x=0,P=\emptyset,R=\{1,\dots,k\},\vec w=A_{I}^\top \K(\MB{Y},\MC{Y})^\top$} 
		\KwProc{}
		\Indp
		$j=\arg \max_{i\in R} (w_i)$\;
		$P=P\cup \{j\}$, $R=R \backslash \{j\}$\;
		$\vec s^P=[(\MB{A}_{I}^P)^\top \K(\MC{Y},\MC{Y}) \MB{A}_{I}^P]^{-1} (\MB{A}_{I}^P)^\top \K(\MB{Y},\MC{Y})^\top$\;
		\textbf{If }$\min(\vec s^P) < 0$ \textbf{then}\\ \label{line:sim}
		\qquad	$\alpha=-\min_{i\in P} [x_i/(x_i-s_i)]$\;
		\qquad  $\vec x:=\vec x+\alpha (\vec s-\vec x)$\;
		\qquad  update $R$ and $P$\;
		\qquad  $\vec s^P=[(\MB{A}_{I}^P)^\top \K(\MC{Y},\MC{Y}) \MB{A}_{I}^P]^{-1} (\MB{A}_{I}^P)^\top \K(\MB{Y},\MC{Y})^\top$\;
		\qquad  $\vec s^R=0$\;
		$\vec x=\vec s$\;	
		$\vec w=\MB{A}_{I}^\top [\K(\MB{Y},\MC{Y})^\top-\K(\MC{Y},\MC{Y}) \MB{A}_{I} \vec x]$\;
		Stop if $R=\emptyset$\;
	\caption{The K-NNLS algorithm}
	\label{fig:knnls}
	\end{algorithm}


\subsection{Non-negative Dictionary Update}\label{sec:KNMF}
As the second part of our algorithm, we want to find the best dictionary $\Phi({\MC{Y}}){\mathbf A}$ which minimizes (Eq.\ref{eq:NNkksvd}) while using the obtained coefficients $\mathbf X$ as the output of NN-KOMP in the previous section. Based on \cite{VanNguyen2013}, the error function $\| \Phi({\MC{Y}})-\Phi({\MC{Y}}){\mathbf A}\MB{X}\|_F^2$ can be reformulated as:
\begin{equation}
\label{eq:e}
\| \Phi({\MC{Y}}) \MB{E}_j-\Phi({\MC{Y}}) \vec a_j \vec x^j\|_F^2 ~; \quad \MB{E}_j=(I-\underset{i\neq j}\sum \vec a_i \vec x^i) 
\end{equation}
$\Phi(\MB{Y}) \MB{E}_j$ is the reconstruction error using all the dictionary columns except $\vec a_j$, and $\vec x^j$ is the corresponding coefficients in the $j$th row of $\MB{X}$ which were estimated by NN-KMOP. Therefore, the dictionary can be updated through solving the (Eq.\ref{eq:e}) for each $\vec a_j$. 
As an important constraint we have to take into account that the optimal dictionary should be used along with non-negative coefficients $\MB{X}$. Accordingly we formulate (Eq.\ref{eq:e}) as the following alternating optimization set:
\begin{alignat}{2}
\label{eq:knnls2}
&\underset{\vec x^j}{\mathrm{min}} \| \Phi({\MC{Y}})\MB{E}_j-\Phi({\MC{Y}}){\vec a_j}\vec x^j\|_F^2 ~~\mathrm{s.t} ~~ \vec x^j \ge 0 \\q
&\underset{\vec a_j}{\mathrm{min}}  \| \Phi({\MC{Y}})\MB{E}_j-\Phi({\MC{Y}}){\vec a_j}\vec x^j\|_F^2 +\|\vec a_j\|_1 ~~\mathrm{s.t} ~~ \vec a_j \ge 0
\label{eq:kista}
\end{alignat}
In order to solve (Eq.\ref{eq:knnls2}), we used the large-scale NNLS algorithm from \cite{Mark2004} which can be easily extended to the kernel version that fits to (Eq.\ref{eq:knnls2}).
\subsubsection{NN-Kernel FISTA:}
In order to solve the optimization problem in (Eq.\ref{eq:kista}), we devised the non-negative kernel FISTA algorithm (NN-K-FISTA) which is a combination of the projected gradient technique \cite{lin2007projected} and the Shrinkage-Thresholding method \cite{Beck2009}. We kernelized \cite{Beck2009}, by calculating $f(\vec a_j)$ and $\nabla f(a)$ based on the Mercer kernel's inner product property as:
\begin{equation}
\begin{array}{l}
\| \Phi(\MC{Y})\MB{E}_j-\Phi(\MC{Y}){\vec a_j}\vec x^j\|_F^2 = \MR{\textbf{tr}}[(\MB{E}_j-\vec a_j \vec x^j)^\top \K(\MC{Y},\MC{Y})(\MB{E}_j-\vec a_j \vec x^j)]	\\
\nabla f(\vec a_j)=-2\K(\MC{Y},\MC{Y})(\MB{E}_j-\vec a_j \vec x^j){\vec{x}^{j\top}} \\
\tau_l(\vec x)=(\vec x-l)(sign(\vec x-l)+1)/2,
\end{array}
\end{equation} 
where $\textbf{tr}()$ denotes the trace operator.
%
As the last step in the dictionary update part, we normalize the dictionary coefficients such that $\| \Phi(\MB{\MC{Y}})\vec a_j\|_2^2=1$.\\ 
To sum up the algorithm, the main loop of the non-negative kernel sparse coding is consist of solving the two main optimization problems (Eq.\ref{eq:knnomp}\&\ref{eq:e}) in a loop until the algorithm's convergence.
	\begin{algorithm}[!t]
	\label{fig:kista}
	\caption{The NNK-FISTA algorithm}
		\SetAlgoLined		
		\KwTask{Solving $\mathrm{min_{\vec a}} f(\vec a)+\lambda \| \vec a \|_1 ~~\mathrm{s.t} ~~ \vec a \ge 0$} 
		\KwInput{function $f(\vec a,\K,\MB{E})$, $\lambda$}
		\KwOutput{non-negative sparse dictionary atom $a$ which fits into (Eq.\ref{eq:kista})}
		\KwInit{$k=0,~t=1,~ 0<\eta<1, ~0<\alpha, ~\delta$} 
		\KwStepk{($k \geq 1$) , find the first possible $i \in \mathbb{N}$ such that with $\alpha_k=\eta^i \alpha_{k-1}$:}
		\Indp
		$\vec a^{k+1}=\tau_{\alpha_k \lambda}(\vec a^k-\alpha_k \nabla f(\vec a^k))$\;
		$f(\vec a^{k+1})-f(\vec a^{k})>(\vec a^{k+1}-\vec a^{k}) \nabla f(\vec a^k) - \|\vec a^{k+1}-\vec a^{k}\|_2^2/(2\alpha)$ \;
		$t_{k+1}=(1+\sqrt{1+4t_k^2})/2$ \;
		Stop if $f(\vec a^{k+1})<\delta$,  otherwise $\vec a^{k+1}=\vec a^{k+1}+(\vec a^{k+1}-\vec a^{k})(t_k-1)/t_{k+1}$			
	\end{algorithm}


\subsection {Label-Consistent NN-KSC Classifier}
In order to use the proposed sparse coding framework as a classifier, we use the idea of ``Label Consistent SC'' from \cite{jiang2013label} which is also kernelized for the K-KSVD algorithm in \cite{chen2015kernel}. Therefore the optimization problem would be reformulated as (Eq.\ref{eq:lcnnksc}), where $\MB{H}$ is the label matrix of training data as 
$\MB{H}(i,j)=1$ if $\MB{Y}_j \in class_i$, and $\MB{Q}$ forces the coefficients $x_i$ to be as similar as possible by defining the columns $\vec q_j=\vec q_i$ if \{$\MB{Y}_j,\MB{Y}_i$\} are in the same class.
\begin{equation}
\begin{array}{ll}
\underset{X,D}{\mathrm{min}}  &\|  \Phi({\mathbf Y})-\Phi({\mathbf Y}){\mathbf A}{\mathbf X}\|_F^2 + \alpha \|\MB{Q}-\MB{Q}\MB{A}\MB{X}\|_F^2 +\beta \|\MB{H}-\MB{H}\MB{A}\MB{X}\|_F^2  +   \|\MB{A}\|_1^2 \\
\mathrm{s.t} &\|\vec x_i\|_0\le T , ~\forall i=1...N. , \quad a_{ij} \ge 0 , \quad \ x_{ij} \ge 0 \\	
\end{array}
\label{eq:lcnnksc}
\end{equation}
Therefore the kernel matrix would be changed to $\widetilde{\K}(\MB{Y}_i,\MB{Y}_i)=\K(\MB{Y}_i,\MB{Y}_j)+\alpha \langle \vec q_i,\vec q_j \rangle+\beta \langle \vec h_i,\vec h_j \rangle$. Using the new $\widetilde{\K}$ as the kernel function, (Eq.\ref{eq:lcnnksc}) can be solved by the proposed NNKSC algorithm. However, the parameters $\alpha$ and $\beta$ should be chosen with a trade-off between the reconstruction error and the classification accuracy.
After optimizing the dictionary matrix $\MB{A}$, in order to classify a validation data $\MB{Y}_{val}$, the NN-KOMP (Eq.\ref{eq:knnomp}) would be used to find sparse code $\vec x_{val}$. Afterward, the label can be determined as 
$$l=\MR{arg min}_j \lvert \MB{1}-\MB{H}(j,:)\MB{A}\vec x_{val} \rvert.$$ 
Since in the supervised setup we have access to the labels of training data, we can take advantage of that to encourage dictionary coefficient vectors $\vec a_j$ to be shaped using just one class of data. As a result, $\vec a_j$ can be assumed as the prototype for one specific class and the subset $A_{I_c}, ~~ I_c\in class_c$ can be used as the class $c$ dictionary and also independent from other classes. To achieve the above, in the NN-K-FISTA algorithm the shrinkage-Threshold should be applied only on the elements of $\vec a_j$ related to input data from classes with lower contributions in $\vec a_j$ via checking $\MB{H}\vec a_j$ between the cycles of (Eq.\ref{eq:knnls2} \& \ref{eq:kista}).

%% file: experiments.tex

\section{Datasets and Experiments}\label{sec:dataset}
In this section, we apply the proposed LC-NNKSC algorithm on the selected experimental setup, and then we compare it with other chosen base-line methods to evaluate its performance. As all the datasets are motion signals, first we use the DTW algorithm to calculate the distance matrix $\mathbf{D}$ and then convert it to the Gram matrix $\mathbf{K}$ using Gaussian kernel $\K (x,y)=exp(-\frac{\| x-y \|^2}{\sigma})$. In order to have a positive semi-definite Gram matrix, we set all its negative eigenvalues to zero (clipping). 
For comparison, we additionally choose the following methods among the kernel based approaches or sparse-coding frameworks:
\\{\textbf{LC-K-KSVD:}}
We use the classification form of Kernel KSVD which was proposed in \cite{chen2015kernel}, and is the closest recent approach to our NNKSC algorithm on account of its structure and its purpose.
\\{\textbf{$k$NN:}} 
We use the $k$-Nearest Neighbor classifier ($k=3$) as a baseline example of a linear approach, with which we classify the data samples based on the pairwise DTW distances between them.
\\{\textbf{Kernel-Kmeans:}}
As another similar kernel-based method, we apply the Kernel K-means clustering \cite{shawe2004kernel} on $\K(\MC{Y},\MC{Y})$ to find $M$ cluster prototypes equal to the size of dictionary $\MB{A}$. Afterward, the distance of each validation data $\MB{Y}_i$ to all prototypes would be calculated as $\vec d_{i}=diag(\MB{E}^\top {\cal K}(\MC{Y},\MC{Y}) \MB{E})-2{\cal K}(\MB{Y}_i,\MC{Y})E+{\cal K}(\MB{Y}_{i},\MB{Y}_{i})$, where $\MB{E}$ is the normalized cluster assignment matrix based on \cite{shawe2004kernel}. After passing $\MB{D}$ into a Gaussian function to convert it to a normalized similarity matrix and keeping the first $T$ biggest elements for each data, the result has a similar structure to $\MB{X}$ in the NNKSC algorithm. Then we feed the coefficients into a multi-class linear SVM to classify the validation data.
\\{\textbf{Affinity Propagation:}} 
We chose the Affinity Propagation algorithm \cite{guan2011text} as an approach which selects prototypes from data samples in a with clustering objective. There, the gram matrix would be used as the similarity matrix, and the class labels of validation data would be determined based on the closest neighboring prototype to each data sample.
\\{\textbf{Kernel PCA:}} 
As the last method for the comparison, we use the kernel-PCA approach from \cite{scholkopf1997kernel} to project DTW based gram matrix $\K(\MC{Y},\MC{Y})$ into $M$ dimension space resulting in the batch of data vectors $X$. Then by using the Eigenvectors related to $\K(\MC{Y},\MC{Y})$, we also calculate $X_{val}$ in the same manner and apply a multi-class linear SVM to classify the generated data vectors.

In order to prevent local optimum cases for each method, we repeat the same experiment with 10 different initial points (or initial dictionaries) and we choose the one with the best result for the comparison. 

\subsection{Evaluation Criteria}\label{sec:eval}
\textbf {Classification:} We measure the correct classification rate as the first metric to evaluate the performance of the algorithms. Each dataset would be randomly split into train, test and the validation parts with 50\%, 25\% and 25\% number of data respectively, and the learning process of the dictionary will be stopped according to the increases in error curve of the test data. Finally, the classification accuracy and other measures would be calculated based on the validation data.
\\
\textbf {Reconstruction error:} Among the utilized methods, only LC-NNKSC and LC-KKSVD are sparse coding frameworks and provide reconstruction error (Eq.\ref{eq:knnomp}) as a measure of their accuracy in sparse representation of the data. 
\\
\textbf {Class-based sparsity:} In addition, because another significant concern of our framework is to provide sparse representation for the data, we also consider the level of sparseness for the coefficients $\MB{X}$. So in order to measure the sparseness in the classification framework, we consider $SP_i=\|\sum\limits_{k \in Class_i} \lvert \vec x_k \rvert\|_0$ as the number of non-zero elements when considering all the sparse codes for each class of data, and we present the best and the worst $SP_i$ for each algorithm.
\\
\textbf {Dictionary sparseness:} Furthermore, to study the dictionary interpretability, we calculate the relevance of each dictionary atom $\vec d_j$ to the data classes. We can find the contribution of each data class in $\vec d_j$ via 
$\vec c=\MB{H}\vec a_j$ where $\MB{H}$ is the class label matrix as in (Eq.\ref{eq:lcnnksc}). Then the dictionary sparseness would be measured as $DS=\max_i (c_i)/{\|\vec c\|_1}$.
%
%
\subsection {Datasets}

\textbf{CMU Motion Dataset}:
We use the Human motion capture dataset from the CMU graphics laboratory \cite{CMU_mocap}, which was captured by Vicon infra-red system.
We combined the movement data of subject 86 from the dataset which is a combination of 9 different types of human movements such as ``walking'',''running'', ``clapping'',... . Then the data is segmented in order to break down the long movements into smaller segments as single periods of each type of motion. Consequently, we obtain 9 classes of data with 10 samples per class, and For implementing LC-NNKSC, we used $\alpha=1$ and $\beta=5$.
%
\\\textbf{Cricket Umpire's Signals:}
%
For our classification experiment we use Cricket Umpire's Signal data provided in 
\cite{ko2005online}. This dataset contains $180$ samples of data from $12$ different classes of umpire signals related to the cricket game. In order to perform the sparse coding classification we choose $\alpha=0.5$ and $\beta=1$.
\\\textbf{Articulatory Words:}
The articulatory words dataset is the facial (ex. lips and tongue) movement signals captured via EMA sensors \cite{wang2014preliminary}. The dataset is used to categorize 25 classes of different words uttered by the subjects in total 575 sample of data. For this dataset we choose $\alpha=0.2$ and $\beta=0.5$.
%
\\\textbf{Squat dataset:}
The squat dataset is gathered in our institute as a part of the large-scale intelligent coaching project. The data is a set of squat movements performed by three coaches while being captured by the optical MOCAP system \cite{waltemate2015realizing}. Each squat is segmented into three movement primitives "preparation", "going down" and "coming up", which generates 87 sample of data and 9 class labels together with the coach labels. Classification of this dataset is performed while using 1 and 0.2 as the $\alpha$ and $\beta$ respectively.
\subsection {Classification Results}
For all the 4 datasets we choose the number of dictionary elements $\mathbf{A_i}$ as twice as the number of total classes. As a rule of thumb, we assume the data in each class can be reconstructed with a low error using only 2 atoms related to that class. We use the same value as the number of prototypes and the mapping dimension in K-Kmeans and K-PCA respectively. Also for the NNKSC and the LC-KKSVD algorithms, we choose the sparsity limit $T=4$ to see how the algorithm is going to use these 2 additional redundancy levels for the dictionary learning and the reconstruction.  \\
\begin{table}
	\renewcommand{\arraystretch}{1.3}
	\caption{Classification accuracy(\%) and the reconstruction error (\%) from applying the selected methods on the chosen datasets}
	\label{tab:Accur_results}
	\centering	
	\begin{tabular}{|c|c|c|c|c|c|c|c|c|} 
			\hline		
		& \multicolumn{2}{c} {CMU} &\multicolumn{2}{|c|} {Cricket Signals} &\multicolumn{2}{|c|} {Articulatory Words} & \multicolumn{2}{|c|} {Squat} \\
		\cline{2-9}		
									    	& Acc   & Rec. Err & Acc & Rec. Err & Acc & Rec. Err & Acc & Rec. Err \\
		\hline		
		LC-NNKSC & 90.91  & 4.17  & 83.33 & 11.07 & 97.33 & 14.52 & 100 & 0.14 \\
			\hline
			LC-KKSVD         & 86.36  & 7.44  & 83.33 & 10.1  & 97.33 & 7.8   & 85  & 3.4 \\
			\hline
			K-Means+SVM     & 68 & -- & 56.25 &-- & 90 & -- & 81 & -- \\
			\hline
			Affinity P.     & 90.1 & -- & 68.75 &-- & 92 & -- & 100 & -- \\
			\hline
			K-PCA+SVM       & 50   & -- & 56.25	& --& 60.66 & -- & 37 & -- \\
			\hline
			kNN             & 86.36 & -- & 79.16 & --& 96.66 & -- & 100 & -- \\
			\hline
	\end{tabular}
\end{table}
In Table.\ref{tab:Accur_results}, the classification result is provided. We can see that For all datasets the proposed algorithm achieved the highest classification accuracy among the evaluated methods; however, for Cricket and Words datasets the LC-KKSVD provided similar accuracy rates to LC-NNKSC (83.33 \% and 97.33 \%) while having smaller reconstruction errors due to the non-negative restrictions. Also in some of the datasets, the affinity propagation and the kNN managed to obtain performance levels equal to the proposed method, for example, both have 100 \% classification accuracy for CMU dataset; nevertheless they do not provide any reconstruction model for the data in comparison to the sparse coding framework.
\\
Table.\ref{tab:sparseness} brings the sparsity analysis of the results, as the best and the worst measures (bDS, wDS) for the relevance of dictionary elements to the classes, as well as the best and worst number of class-based sparsity (bSP, bSP). According to the table.\ref{tab:sparseness}, LC-NNKSC provide models for the datasets with better sparseness regarding both the dictionary atoms and the class data reconstruction. For all datasets, it defines each dictionary atom using the data of a single class which results in almost 100 \% dictionary sparseness. For the squat dataset, the algorithm managed to reconstruct the data of each class using only one specific atom (wSP=bSP=1), meaning that only half of the dictionary is needed to model this data with NNKSC. Also, due to the value of wSP in Cricket and Words data (4 and 3 respectively), apparently there exist classes which require more than 2 dictionary atoms to be reconstructed and categorized efficiently.
\\
The LC-KKSVD too has a high classification accuracy for Cricket and Words data, but this performance is lower than Affinity Propagation in the other 2 datasets. Furthermore, from the sparseness point of view, it is outperformed even by Affinity propagation by providing lower class-based sparsity. 

\begin{table}
	\renewcommand{\arraystretch}{1.8}
	\caption{The best and worst class-based sparseness (bSP and wSP), and the best and worst dictionary sparseness (bDS(\%) and wDS(\%)) for the different selected approaches}
	\label{tab:sparseness}
	\centering
	\resizebox{1\textwidth}{!}{%
	\begin{tabular}{|c||c|c|c|c||c|c|c|c||c|c|c|c||c|c|c|c|} 
	\hline		
		& \multicolumn{4}{c||} {CMU} &\multicolumn{4}{|c||} {Cricket Signals} &\multicolumn{4}{|c||} {Articulatory Words} & \multicolumn{4}{|c|} {Squat dataset} \\
		\cline{2-17}
		 			 					   & bSP & wSP & bDS & wDS & bSP & wSP & bDS & wDS & bSP & wSP & bDS & wDS  & bSP & wSP & bDS & wDS \\
		\hline
		LC-NNKSC &  1  &  2   &  100 &   100  &  1   &  4   &  100   &  100   &  1   & 3  & 100 & 98.1 &  1 & 1  & 100 & 100 \\
			\hline
			LC-KKSVD         &  5  &  9   &  100 &   76   &   5  &  13 &   100  &  44   &  5  &  16  &  100   &  56   &  3	  & 8  & 100 & 87 \\
			\hline
			Affinity P.      &  4  &  6   &  --  &   --   &   6  &  4  &   --   &  --   &  5  &  11  &  --    &  --   &  4	  & 5  & 100 & 87 \\
			\hline
			K-Means      &  4  &  17  &  100  &  50   &  5   &  27   &  100   &  16   &   5  & 50    & 100    &  50  &  4   & 12  & 100 & 60  \\
			\hline
			
	\end{tabular}}
\end{table}

%% file: conclusion.tex
\section{Conclusion}
In this paper, we presented a non-negative kernel based sparse coding approach for modeling and classification of motion data. According to the results, the non-negative approach provides much sparser representation for the data comparing to the conventional Kernel SC method, using a fewer number of prototypes to reconstruct the motion signals. Additionally, where it is possible, the LC-NNKSC approach forces dictionary elements to be created using a positive linear combination of data only from individual classes. 
By this strategy, the obtained dictionary can be easily broken down to class-based dictionaries as separate prototype-based models for each class of data. In addition, these sub-dictionaries can be used as a warm start in further classification tasks even when there is a different combination of classes. Altogether, the LC-NNKSC classifier provides dictionary prototypes and sparse coefficients which are more class-based consistent and makes it possible to have individual models for reconstruction of each class of data as well as for its classification. \\
Based on the strength of this method in constructing prototype based models for the motion data, there is considerable potential for future works on the clustering and designing generative models of motion data using this framework or its variants. 